\documentclass[10pt]{article}
\usepackage[utf8]{inputenc}
\usepackage{hyperref}

\usepackage{arabtex}

\title{A Panoramic Survey of \\ Natural Language Processing in the Arab World}
\author{Kareem Darwish, Nizar Habash,\\
Mourad Abbas, Hend Al-Khalifa, Huseein T. Al-Natsheh,\\
Samhaa R. El-Beltagy, Houda Bouamor, Karim Bouzoubaa,\\
Violetta Cavalli-Sforza, Wassim El-Hajj,\\
Mustafa Jarrar, and Hamdy Mubarak}

\date{October 2020}
 
\begin{document}

\maketitle
\setarab
\vocalize

\section{Natural Language and its Processing} % Round 2 - DONE 

The term {\it natural language} refers to any system of symbolic communication (spoken, signed or written) that has evolved naturally in humans without intentional human planning and design.  This distinguishes natural languages such as Arabic and Japanese from artificially constructed languages such as Esperanto or Python.   Natural language processing (NLP), also called computational linguistics or human language technologies, is the sub-field of artificial intelligence (AI) focused on modeling natural languages to build applications such as speech recognition and synthesis, machine translation, optical character recognition (OCR), sentiment analysis (SA), question answering, dialogue systems, etc.  NLP is a highly interdisciplinary field with connections to  computer science, linguistics, cognitive science, psychology, mathematics and others.

Some of the earliest AI applications were in NLP (e.g., machine translation); and the last decade (2010-2020) in particular has witnessed an incredible increase in quality, matched with a rise in public awareness, use, and expectations of what may have seemed like science fiction in the past.  NLP researchers pride themselves on developing language independent models and tools that can be applied to all human languages, e.g. machine translation systems can be built for a variety of languages using the same basic mechanisms and models.  However, the reality is that some languages do get more attention (e.g., English and Chinese) than others (e.g., Hindi and Swahili).  Arabic, the primary language of the Arab world and the religious language of millions of non-Arab Muslims is somewhere in the middle of this continuum.  Though Arabic NLP has many challenges, it has seen many successes and developments.

Next we discuss Arabic's main challenges as a necessary background, and we present a brief history of Arabic NLP. We then survey a number of its research areas, and close with a critical discussion of the future of Arabic NLP.  % An extended version of this article including almost 200 citations and links is on Arxiv.\footnote{\url{https://arxiv.org/pdf/XXXXXXXXX}}

\section{Arabic and its Challenges} % Round 2 - DONE 

Arabic today poses a number of modeling challenges for NLP: morphological richness, orthographic ambiguity, dialectal variations, orthographic noise, and resource poverty.
We do not include issues of right-to-left Arabic typography, which is an effectively solved problem (although not universally implemented).

\paragraph{Morphological Richness}
Arabic words have numerous forms resulting from a rich inflectional system that includes features for gender, number, person, aspect, mood, case, and a number of attachable clitics.  As a result, it is not uncommon to find single Arabic words that translate into five-word English sentences: <wasayadrusuwnahA> {\it wa+sa+ya-drus-uuna+ha} `and they will study it'.  This challenge leads to a higher number of unique vocabulary types compared to English, which is challenging for machine learning models.

\paragraph{Orthographic Ambiguity}
The Arabic script uses optional diacritical marks to represent short vowels and other phonological information that is important to distinguish words from each other. These marks are almost never used outside of religious texts and children's literature, which leads to a high degree of ambiguity.  Educated Arabs do not usually have a problem with  reading undiacritized Arabic, but it is a challenge for Arabic learners and computers.
This out-of-context ambiguity in Standard Arabic leads to a staggering 12 analyses per word on average: e.g. the readings of the word <ktbt> {\it ktbt} (no diacritics) includes <katab"tu> {\it katabtu} `I wrote', 
%<katabta> `you [male] wrote',
%<katabti> ` you [female] wrote',
<katabat"> {\it katabat} `she wrote', and the quite semantically distant
<katibat> {\it ka+tibat} `such as Tibet'.

\paragraph{Dialectal Variation}
Arabic is also not a single {\it language} but rather a family of historically linked varieties, among which Standard Arabic is the official language of governance, education and the media, while the other varieties, so-called dialects, are the languages of daily use in spoken, and increasingly written, form. Arab children grow up learning their native dialects, such as  Egyptian, Levantine, Gulf, or Moroccan Arabic, which have their own grammars and lexicons that differ from each other and from Standard Arabic. For example, the word for `car' is <sy"ArT> {\it sayyaara} in Standard Arabic, <`rbyT> {\it arabiyya} in Egyptian Arabic, <krhbT> {\it karhba} in Tunisian Arabic, and <mwtr> {\it motar} in Gulf Arabic.
The differences can be significant to the point that using Standard Arabic tools on dialectal Arabic leads to quite sub-optimal performance.  
For instance, \cite{Khalifa:2016:large} report that using a state-of-the-art tool for MSA morphological disambiguation on Gulf Arabic returns POS tag and lemma accuracy at about 72\% and 64\% respectively, compared to the performance on MSA, which is 96\% for both \cite{pasha2014madamira}. 

\paragraph{Orthographic Inconsistency}

Standard and dialectal Arabic are  both written with a high degree of spelling inconsistency, especially on social media:
%\kareem{For dialects, they lack spelling standards that are widely adopted. As for MSA,}
 %32\% 
A third of all words in MSA comments online have spelling errors \cite{Zaghouani:2014:large}; 
 and dialectal Arabic has no official spelling standards, although there are efforts to develop such standards computationally, such as the work on CODA, or Conventional Orthography for Dialectal Arabic \cite{Habash:2018:unified}. 
Furthermore, Arabic can be encountered online written in other scripts, most notably, a romanization called Arabizi that attempts to capture the phonology of the words \cite{Darwish:2014:arabizi}.

\paragraph{Resource Poverty}
Data is the bottleneck of NLP: this is true for rule-based approaches that need lexicons and carefully created rules; and for Machine Learning (ML) approaches that need corpora and annotated corpora.
Although Arabic unannotated text corpora are quite plentiful, Arabic morphological analyzers and lexicons  as well as annotated and parallel data in non-news genre and in dialects are limited.

None of the above-mentioned issues are unique to Arabic - e.g. Turkish and Finnish are morphologically rich; Hebrew is orthographically ambiguous; and many languages have dialectal variants.  However, the combination and degree of these phenomena in Arabic creates a particularly challenging situation for NLP research and development.
For more information on Arabic computational processing challenges, see \cite{farghaly2009arabic,Habash:2010:introduction}.

\section{A Brief History of NLP in the Arab World}  % Done Round 2

Historically, Arabic NLP can be said to have gone through three waves. The first wave was in the early 1980's with the introduction of Microsoft MS-DOS 3.3 with Arabic language support.
% \cite{rippin1991arabic}. 
In 1985 the first Arabic morphological analyzer was developed by Sakhr. Most of the research in that period focused on morphological analysis of Arabic text and using rule-based approaches. 
Sakhr has also continued leading research and development in Arabic computational linguistics by developing the first syntactic and semantic analyzer in 1992 and Arabic optical character recognition in 1995. Sakhr  also produced many commercial products and solutions including Arabic to English machine translation, Arabic text-to-speech (TTS), and an Arabic search engine.  This period almost exclusively focused on Standard Arabic with a few exceptions related to work on speech recognition \cite{Gadalla:1997:callhome}.

The second wave was during the years 2000-2010. Arabic NLP gained increasing importance in the Western world especially after September 11.  
The USA funded large projects for companies and research centers to develop  NLP tools for Arabic and its dialects including: machine translation, speech synthesis and recognition, information retrieval and extraction, text to speech, and named entity recognition \cite{farghaly2009arabic,Habash:2010:introduction,Olive:2011:handbook}.  
Most of the  systems developed in that period  used machine learning, which was on the rise in the field of NLP as a whole. In principle, ML required far less linguistic knowledge than rule-based approaches, and was fast and more accurate. However it needed a lot of data, some of which was not easy to collect, e.g. dialectal Arabic to English parallel texts. Arabic's rich morphology exacerbated the data dependence further. So, this period saw some successful instances of hybrid systems that combine rule-based morphological analyzers with ML disambiguation which relied on the then newly created Penn Arabic Treebank (PATB) \cite{Habash:2005:MADA,Maamouri:2004:patb}.  
The leading universities, companies and consortia at the time were: Columbia University, University of Maryland, IBM, BBN, SRI, the Linguistic Data Consortium (LDC), and European Language Resources Association (ELRA).
%consortium was established 2008 for cooperation between Arabic and European Union countries for developing Arabic language resources.

The third wave started in 2010, when the research focus on Arabic NLP came back to the Arab world. This period witnessed a proliferation of Arab researchers and graduate students interested in Arabic NLP and an increase in publications in top conferences from the Arab world. Active universities include New York University Abu Dhabi (NYUAD), American University in Beirut (AUB), Carnegie Mellon University in Qatar (CMUQ), King Saud University (KSU), Birzeit University (BZU), Cairo University, etc. Active research centers include Qatar Computing Research Institute (QCRI), King Abdulaziz City for Science and Technology (KACST), etc. It should be noted that there are many actively contributing researchers in smaller groups across the Arab world.
This period also overlapped with two major independent developments: one is the rise of deep learning and neural models and the other is the rise of social media.  The first development affected the direction of research pushing it further into the ML space; and the second led to the increase in social media data, which introduced many new challenge at a larger scale: more dialects and more noise. 
This period also witnessed a welcome increase in Arabic language resources and processing tools, and a heightened awareness of the importance of AI for the future of the region -- e.g. the UAE now has a ministry for AI specifically. 
Finally, new young and ambitious companies such as Mawdoo3 are competing for a growing market and expectations in the Arab world.  
It's important to note that throughout the different waves, researchers of different backgrounds (Arab and non-Arab) contributed and continue to contribute to the developments in the Arabic NLP.

\section{Arabic Tools and Resources} % Round 2 done -- and restructured to address comments

We organize this section on Arabic tools and resources into two parts: first, we discuss enabling 
technologies which are the basic resources and utilities that are not 
user-facing products; and, second, 
we discuss a number of advanced user-targeting applications.

\subsection{Arabic Enabling Resources and Technologies}

\subsubsection{Corpora and Lexical Resources} %Enabling resources.

Resource construction is a lengthy and costly task that requires significant teamwork among linguists, lexicographers, and publishers over an extended period of time.  

\paragraph{Corpora} NLP relies heavily on the existence of corpora for developing and evaluating its models, and the performance of NLP applications directly depends on the quality of these corpora. 
Historically, Al-Khalil ibn Ahmad Al-Farahidi, was among the first to assemble a large collection of texts in the 8th century \cite{versteegh2014arabic}. 
Textual corpora are classified at a high level as annotated and unannotated corpora. 

Annotated corpora are naturally a subset of unannotated corpora that have been enriched with additional information such as contextual morphological analyses, lemmas, diacritizations,  part-of-speech tags,  syntactic analyses, dialect IDs, named entities, sentiment, and even parallel translations.  The more information, the costlier the process is to create such corpora.  %Annotated corpora are the core of supervised machine learning NLP approaches. 
For Arabic, the main collections of annotated corpora were created in its second wave, mostly outside the Arab world. 
the most notable annotated resource is the LDC's Penn Arabic Treebank (PATB) \cite{Maamouri:2004:patb}, 
which provides a relatively large MSA corpus that is morphologically analyzed, segmented, lemmatized, tagged with fine-grained parts of speech, diacritized, and parsed. PATB has enabled much of the Arabic NLP research since its creation. 
The Prague Arabic Dependency Treebank (PADT) was the first dependency representation treebank for Arabic \cite{Smrz:2002:prague}. 
The Columbia Arabic Treebank (CATiB) was an effort to develop a simplified dependency representation with a faster annotation scheme for MSA \cite{Habash:2009:catib,Taji:2018:arabic}.
The University of Leeds' Quranic Arabic Corpus is a beautifully constructed treebank that uses traditional morpho-syntactic analyses of the Holy Quran \cite{Dukes:2010:dependency,Dukes:2010:morphological}.
With the rising interest in dialectal data, there have been many efforts to collect and annotate dialectal data \cite{al2012yadac,Al-Twairesh:2018:suar,Bouamor:2014:multidialectal,Bouamor:2018:madar,diab2010colaba,el2020habibi,Gadalla:1997:callhome,sadat-kazemi-farzindar:2014:SocialNLP,Salama:2014:youdacc,Smaili:2014:building}. 
The LDC was first to create a Levantine and an Egyptian Arabic Treebanks \cite{Maamouri:2014:developing}.

In the Arab world, the efforts are relatively limited in terms of creating annotated corpora. 
Examples include BZU's Curras, the Palestinian Arabic  annotated corpus \cite{Jarrar:2016:curras}, 
 NYUAD's Gumar, the Emirati Arabic annotated corpus \cite{Khalifa:2018:morphologically},  
and Al-Mus’haf Quranic Arabic corpus \cite{zeroual2016}, which  is  based on the Quran and includes morpho-syntactic annotations such as stems, stem patterns, and roots.
Another annotation effort with a focus on MSA spelling and grammar correction is the Qatar Arabic Language Bank (QALB), which was developed between Columbia  and CMUQ \cite{Zaghouani:2014:large}.
Other  specialized annotated corpora developed in the Arab world include NYUAD's parallel gender corpus with sentences in masculine and feminine for anti-gender bias research,  the Arab-Acquis corpus pairing Arabic with all of Europe's languages for a portion of European parliamentary proceedings, and the MADAR corpus of parallel dialects created in collaboration with CMUQ.

In contrast to annotated corpora, there are many unannotated datasets.
Most large data sets also started outside the Arab world, e.g.,
the Agence France Press document collection, which is heavily used for Arabic information retrieval evaluation, the LDC's Arabic Gigaword \cite{Parker:2011:arabic}, 
Arabic Wikipedia, and the ArTenTen corpus \cite{arts2014artenten}. % and 
% Shamela, a large-scale corpus (1B words) covering the past 14 centuries of Arabic.
%\cite{belinkov-etal-2016-shamela}, 
%
Important collections in the Arab World include: the
%https://www.aclweb.org/anthology/W14-3602.pdf ICA /Ansary
International Corpus of Arabic (Bibliotheca Alexandrina); Shamela, a large-scale corpus (1B words) covering the past 14 centuries of Arabic\footnote{\url{https://shamela.ws/}};
the Tashkeela corpus containing 75M fully vocalized words (National Computer Science Engineering School in Algeria) \cite{zerrouki17};   
NYUAD's Gumar Gulf Arabic corpus containing over 100M words of internet novels \cite{Khalifa:2016:large}; 
and Abu El-Khair corpus (Umm Al-Qura University, KSA) \cite{el20161}.  
The success of word embedding  models \cite{Mikolov2013,Pennington2014}  
trained on unannotated data  and resulting in improved performance for NLP tasks with little or no feature engineering have led to many contributions in Arabic NLP \cite{al2015deep,farha2019mazajak,Soliman2017}.
The more recent appearance of contextualized embeddings trained on unannotated data, such as BERT \cite{devlin2018bert}, is creating promising possibilities for improving many Arabic NLP tasks. At the time of writing this article, a handful of contextualized embedding models are known to support Arabic including Multilingual BERT \cite{bert}, Arabic BERT \cite{safaya2020kuisail}, 
AraBERT (AUB) \cite{antoun2020arabert}, GigaBert (Ohio State U.) \cite{lan2020gigabert}, Marbert (U. of British Columbia) \cite{abdulmageed2020microdialect}, and QARiB (QCRI).

%Multilingual BERT, %\cite{bert}, 
%Arabic BERT, %\cite{safaya2020kuisail},  
%AraBERT \cite{antoun2020arabert}, 
%GigaBert \cite{lan2020gigabert}, 
%and MARBert \cite{abdulmageed2020microdialect}.
%Marbert does.   However, none of these models has wide Arabic dialect coverage. 

\paragraph{Lexical Resources} % ADD CITATIONS HERE
We can distinguish three types of lexical resources (aka, lexicons, dictionaries and databases): (a) morphological resources that encode all inflected forms of words, (b) lexical resources that are lemma based, such as machine readable monolingual and multilingual dictionaries, and (c) semantic resources that link lemmas to each other, such as wordnets and ontologies.  These resources are useful for a variety of NLP tasks.

Some of the earliest publicly available Arabic lexical resources were created outside of the Arab world in the second wave mentioned earlier. The Buckwalter Arabic Morphological Analyzer (BAMA), with its extended version called Standard Arabic Morphological Analyzer (SAMA), both available from the LDC, provided one of the first stem databases with tags and morphological solutions, and are used in a number of tools. Elixir-FM is a functional morphology analyzer developed in Charles University in Czech Republic \cite{Smrz:2007:elixirfm}. %cite Smrz
The DIINAR \cite{dichy2005dinar} %(Dichy cite)
Arabic morphological database is a full form resource developed in France.
%
%\cite{Graff09}.  
The Tharwa lemma-based lexicon was developed at Columbia University and included 70k entries in Egyptian Arabic, MSA and English; and later extended with Levantine Arabic \cite{Diab:2014:tharwa}.
Arabic WordNet  is a semantic lexicon consisting of about 11k synsets with subset and superset relationships between concepts and linked to a number of other languages through the Global Wordnet effort. 
This effort was done by a number of American and European universities \cite{awn06}.
And the Arabic VerbNet \cite{Mousser2013} 
classifies verbs that have the same syntactic descriptions and argument structure (U. of Konstanz, Germany).

Some of the efforts in the Arab world led to multiple notable resources.
Al-Khalil morphological analyzer is a large morphological database for Arabic developed by researchers in Morocco and Qatar \cite{Boudchiche:2017:alkhalil}
Calima Star is an extension of the BAMA/SAMA family done at NYUAD and is part of the CAMeL Tools toolkit \cite{Obeid:2020:cameltools,Taji:2018:arabic-morphological}. %cite calima* and cameltools
%
%Jarrar and Amayreh 
BZU developed a large Arabic lexicographic database constructed from 150 lexicons that are diacritized and standardized \cite{JA19}. 
The MADAR project (NYUAD and CMUQ) includes a lexicon with 47k lemma entries covering 25 city dialects in Parallel \cite{Bouamor:2018:madar}.
Other lexicons have been developed for 
%???
%\cite{namly20} developed one of the most comprehensive lexicons with more than 7M stems and corresponding lemmas. 
Algerian \cite{abidi2018automatic}, 
Tunisian \cite{sghaier2017tunisian}, 
and Morrocan \cite{tachicart2014building}.  
Finally, in terms of semantic lexical resources, the BZU Arabic Ontology \cite{jarrarAraOnto} 
is a formal Arabic wordnet with more than 20k concepts that was built with ontological analysis in mind and is linked to the Arabic Lexicographic Database,  Wikidata, and other resources \cite{JAM19}.

More Arabic resources can be found in known international repositories (namely ELRA/ELDA, LDC, and CLARIN) or directly from their authors’ websites \cite{farghaly2009arabic,Habash:2010:introduction,Zaghouani:2014:critical}.
%zeroual2018thesis}. %Zaghouani:2014:critical, 
Unfortunately, many are not interoperable, have been built using different tools and assumptions, released under propriety licenses, and a few are comprehensive. Serious, well-planned, and well-coordinated investment in resources will be instrumental for the future of Arabic NLP. 

%%%%%%%%%%%%%%%%%

\subsubsection{Morphological Processing} %Round 2 done!

Given the  challenges of Arabic morphological richness and ambiguity, morphological processing has received a lot of attention.
The task of morphological analysis refers to the generation of all possible readings of a particular undiacritized word out of context. Morphological disambiguation is about identifying the correct in-context reading. This broad definition allows us to think of word-level tasks such as  Part-of-Speech (POS) tagging, stemming, diacritization and tokenization as sub-types of morphological disambiguation that focus on specific aspects of ambiguity.

Most work on Arabic morphological analysis and  disambiguation is on MSA; however there is a growing number of efforts on dialectal Arabic \cite{Darwish:2018:multi-dialect,khalifa-etal-2020-morphological,pasha2014madamira,Samih:2017:neural,Zalmout:2018:noise-robust}.
There are a number of commonly used morphological analyzers for Standard and dialectal Arabic (Egyptian and Gulf), e.g., BAMA, SAMA, Elixir-FM, Al-Khalil, Calima Egyptian, and CAlima Star \cite{Boudchiche:2017:alkhalil,Buckwalter:2002:buckwalter,Graff:2009:standard,Habash:2012:morphological,Khalifa:2017:morphological,Smrz:2007:elixirfm}.
Some of the morphological disambiguation systems disambiguate the analyses that are produced by a morphological analyzer using PATB as a training corpus, e.g.,  MADAMIRA (initially developed at Columbia U.), and other variants of it from NYUAD \cite{Khalifa:2016:yamama,Obeid:2020:cameltools,pasha2014madamira,Zalmout:2017:dont}.
Farasa (from QCRI) uses independent models for tokenization \cite{Abdelali:2016:farasa} 
and POS tagging \cite{darwish2017arabic}.
%Khoja:2001:apt

%

\subsubsection{Syntactic Processing}  %Round 2 done!
% Hend Al-Khalifa 
 
Syntactic parsing is the process of generating a parse tree representation for a sentence that indicates the relationship among its words. For example, a syntactic parse of the sentence <qra'a Al.tAlbu AlktAba Aljdyda> `[lit.] {\it read the-student the-book the-new}; the student read the new book' would indicate that the adjective {\it the-new} modifies the noun {\it the-book}, which itself is the direct object of the verb {\it read}.

There are many syntactic representations. Most commonly used in Arabic are the PATB constituency representation \cite{Maamouri:2004:patb}, 
the CATiB dependency representation \cite{Habash:2009:catib}, 
and the Universal Dependency (UD) representation \cite{Taji:2017:universal}.
All of these were developed outside of the Arab world. The UD representation is an international effort, where NYUAD is the representative of the Arab world on Arabic.

The most popular  syntactic parsers for Arabic are: Stanford, Farasa (QCRI), and CamelParser (NYUAD).  Stanford is a statistical parser from the Stanford Natural Language Processing Group that can parse English, German, Arabic and Chinese.  For Arabic, it uses a probabilistic context free grammar that was developed based on PATB \cite{Green:2010:better}. Updated versions of the parser use a shift-reduce algorithm \cite{zhu2013fast} and neural transition-based dependency parsing \cite{chen2014fast}. 
Farasa is an Arabic NLP toolkit that provides syntactic constituency and dependency parsing \cite{zhang2015randomized}.
CamelParser is a dependency parser trained on CATiB treebank using MaltParser \cite{Nivre:2006:maltparser}, 
a language-independent and data-driven dependency parser \cite{Shahrour:2016:camelparser}. 
A discussion and survey of some of the Arabic parsing work is presented in \cite{Habash:2010:introduction}.

\subsubsection{Named Entity Recognition} %Round 2 done!
% Samhaa El-Beltagy
Named Entity recognition (NER) is the task of identifying one or more consecutive words in text that refer to objects that exist in the real-world (named entities), such as organizations, persons, locations, brands, products, foods, etc.  NER is essential for extracting structured data from an unstructured text, relationship extraction, ontology population, classification, machine translation, question answering, and other applications. 
Among the challenges facing Arabic NER compared to English NER is the lack of letter casing which strongly helps English NER and the high degree of ambiguity, including especially confusable proper names and adjectives, e.g. <krym> {\it kariym} can be the name `Kareem' or the adjective `generous'. 
% Unlike English, where named entities, such as person names, are capitalized, Arabic lacks casing, which complicates Arabic NER. Further, Arabic named entities often correspond to Arabic adjectives or nouns. For example, the Arabic names 'Jameela' and 'Adel' are also the common adjectives `beautiful' and `just'.  This inherent ambiguity means that matching entries in named entity gazetteers is highly error prone. 

Arabic NER approaches include the use of hand-crafted heuristics, machine learning, and hybrids of both with heavy reliance on gazetteers \cite{Benajiba:2008:arabic,darwish2013named,Shaalan:2014:survey}. Recent approaches exploited cross-lingual links between Arabic and English knowledge bases to expand Arabic gazetteers and to carry over capitalization from English to Arabic \cite{darwish2013named}. 
Much of the earlier work on Arabic NER focused on formal text, typically written in MSA. However, applying models trained on MSA text to social media (mostly dialectal) text has led to unsatisfactory results \cite{darwish2014simple}. 
%The aforementioned Arabic NER challenges suggest that understanding the context in which a word appears can address linguistic ambiguity and significantly improve NER.  
Recent contextualized embeddings and other deep learning approaches such as sequence to sequence models and convolutional neural networks have led to improved results for Arabic NER \cite{ali2019boosting,DBLP:journals/csl/KhalifaS19,liu-etal-2019-arabic,Obeid:2020:cameltools}. 
It is expected that the use of contextualized embeddings trained on larger corpora of varying Arabic dialects, coupled with the use of deep learning models is likely to contribute positively to Arabic NER. 
As with other utilities, early research was done outside of the Arab world, but more work is now happening in the Arab world.
A extensive list of Arabic NER challenges and solutions can be found in \cite{Shaalan:2014:survey}.

\subsubsection{Dialect Identification} %Round 2 done
% Houda Bouamor
% Dialect id/context switching

Dialect identification (DID) is the task of automatically identifying the dialect of a particular segment of speech or text of any size (i.e., word, sentence, or document). This task has been attracting increasing attention in NLP for a number of language varieties \cite{zampieri-EtAl:2018:VarDial}. 
DID has been shown to be important for several NLP tasks where prior knowledge about the dialect of an input text can be helpful, such as % machine translation, sentiment analysis, and author profiling.
machine translation~\cite{salloum-EtAl:2014:P14-2}, sentiment analysis~\cite{altwairesh-alkhalifa-alsalman:2016:P16-1}, and author profiling~\cite{sadat-kazemi-farzindar:2014:SocialNLP}.

Early Arabic multi-dialectal data sets and models focused on the regional level \cite{Biadsy:2009:using,Bouamor:2014:multidialectal,darwish-sajjad-mubarak:2014:EMNLP2014,Elfardy:2014:aida,Guellil:2016,Meftouh:2015:machine,zaidan2011arabic}.
The Multi Arabic Dialects Application and Resources (MADAR) project aimed to create a finer grained dialectal corpus and lexicon \cite{Bouamor:2018:madar}. 
The data was used for dialectal identification at the city level \cite{obeid-etal-2019-adida,Salameh:2018:fine-grained} 
of 25 Arab cities, and was used in a shared task for DID \cite{bouamor2019madar}.  
The main issue with that data is that it was commissioned and not naturally occurring.  Concurrently, larger Twitter-based datasets covering 10-21 countries were also introduced  \cite{abdelali2020Arabic,Abdul-Mageed:2018:you,Mubarak:2014:using,Zaghouani:2018:araptweet}. 
The Nuanced Arabic Dialect Identification (NADI) Shared Task \cite{mageed-etal-2020-nadi} 
followed earlier pioneering works by providing country-level dialect data for 21 Arab countries, and introduced a province-level identification task aiming at exploring a total of 100 provinces across these countries.
Also here, earlier efforts started in the west, most notably work in Johns Hopkins University \cite{Zaidan:2013:arabic}, but more work is happening now in the Arab world (NYUAD and QCRI) \cite{abdelali2020Arabic,bouamor2019madar}.

\subsubsection{Infrastructure} %Round 2 done
% Karim Bouzoubaa

To aid the development of NLP systems, a number of multi-lingual infrastructure toolkits have been developed, e.g., GATE\footnote{\url{https://gate.ac.uk}}, Stanford CoreNLP\footnote{\url{https://stanfordnlp.github.io/CoreNLP/}} and UIMA\footnote{\url{https://uima.apache.org/d/uimaj-current/}}. Their philosophy is to gather and develop several NLP tools within a single and homogeneous structure that is flexible, extensible and modular. 
They offer researchers easy access to several tools through command-line interfaces (CLIs) and application programming interfaces (APIs), thus eliminating the need to develop them from scratch every time.
While Arabic NLP has made significant progress with the development of several enabling tools, such as POS taggers, morphological analyzers, text classifiers, and syntactic parsers, there is a limited number of homogeneous and flexible Arabic infrastructure toolkits that gather these components. 
MADAMIRA \cite{pasha2014madamira} is a Java-based system providing  solutions to fundamental NLP tasks for Standard and Egyptian Arabic. These tasks include diacritization, lemmatization, morphological analysis and disambiguation, POS tagging, stemming, glossing,  (configurable) tokenization, base-phrase chunking and NER.\footnote{\url{https://camel.abudhabi.nyu.edu/madamira/}}  MADAMIRA's signature approach is to address multiple morphology related tasks together in one fell swoop producing deep linguistic representations.
Farasa\footnote{\url{http://qatsdemo.cloudapp.net/farasa/}} is a collection of Java libraries and CLIs for MSA. These include separate tools for diacritization \cite{darwish2020arabic,mubarak2019highly}, segmentation \cite{Abdelali:2016:farasa}, lemmatization \cite{mubarak2018build}, POS tagging \cite{darwish2017arabic}, parsing \cite{zhang2015randomized}, and NER \cite{Darwish:2013:Named}. 
%This allows for a more flexible use of components compared to MADAMIRA. 
%
SAFAR\footnote{\url{http://arabic.emi.ac.ma/safar/}} is a Java-based framework bringing together all layers of Arabic NLP: resources, pre-processing, morphology, syntax, and semantics. 
%SAFAR's goal is to gather, within a single homogeneous infrastructure, the already developed and available Arabic tools to facilitate access to them, to enable comparing them.
%
CAMeL Tools \cite{Obeid:2020:cameltools} is a recently developed collection of open-source tools, developed in Python, that supports both MSA and Arabic dialects.\footnote{\url{https://github.com/CAMeL-Lab}}  It currently provides APIs and CLIs for pre-processing, morphological modeling, dialect identification, NER, and sentiment analysis. 
Other notable efforts include AraNLP \cite{althobaiti2014aranlp}, 
ArabiTools,\footnote{\url{https://www.arabitools.com/}} and 
Adawat.\footnote{\url{http://adawat.sourceforge.net/}}  
A feature comparison of some Arabic infrastructures can be found in \cite{Obeid:2020:cameltools} while a detailed survey and a software engineering comparative study can be found in \cite{jaafar2018survey}. Whatever the specific choice a researcher might make, the trend to use infrastructures more extensively will significantly save time and change the way Arabic applications are designed and developed.

%DEPRECATED SINCE 2018 ATKS\footnote{\url{https://www.microsoft.com/en-us/research/project/arabic-toolkit-service-atks/}}, 

 %%%%%%%%%%%%%%%%
\subsection{Arabic NLP Applications}

\subsubsection{Machine Translation} % Roudn 2 done

Machine Translation (MT) is one of the earliest and most worked on areas in NLP.
The task is to map input text in a source language such as English to an output text in a target language such as Arabic.
Early MT research was heavily rule-based \cite{guessoum2001methodology,hatem2010syntactic,ibrahim1991MA,rafea1992mutargem,shaalan2004machine}; 
however now it is almost completely corpus-based using a range of statistical and deep learning models, depending on resource availability.

For MSA, parallel data in the news domain is plentiful.\footnote{Linguistic Data Consortium (LDC) resources: LDC2004T18, LDC2004T14, and LDC2007T08.} There are other large Arabic parallel collections under the OPUS project \cite{tiedemann2016parallel} 
and as part of the UN corpus \cite{Rafalovitch:2009:united}.  
Other specialized corpora include the Arab-Acquis corpus pairing with European languages (NYUAD) \cite{Habash:2017:parallel} 
and the AMARA educational domain parallel corpus (QCRI) \cite{Abdelali:2014:amara}.
Dialectal parallel data are harder to come by and most are commissioned translations \cite{Bouamor:2014:multidialectal,Bouamor:2018:madar,Erdmann:2017:low,Meftouh:2015:machine,Zbib:2012:machine}.

There are many other efforts in Statistical MT (SMT) from and to Arabic \cite{carpuat2010improving,ElKholy:2010:orthographic,elming2009syntactic,Habash:2006:arabic-preprocessing,hasan2006creating,hatem2008modified,nguyen2008context,tahir2010knowledge}.
Recently, Deep Neural Networks have been adopted for Arabic Machine Translation as in \cite{al2020neural,2016arXiv160602680A,DBLP:journals/mt/AmeurGM19,gashaw2019amharicarabic,oudah-etal-2019-impact}. 
While most researched MT systems for Arabic target English, 
there have been efforts on  MT for Arabic and other languages, e.g., Chinese \cite{habash2009improving}, 
Russian \cite{Zalmout:2017:optimizing}, 
Japanese \cite{Inoue:2018:parallel}, 
and all of the  European Union languages \cite{Habash:2017:parallel}. 

MT for Arabic dialects is more difficult due to limited resources, but there are noteworthy efforts exploiting similarities between MSA and dialects in universities and research group around the world  \cite{Sajjad:2013:translating,sajjad2016egyptian,Salloum:2011:dialectal,Sawaf:2010:arabic,Shaalan:2007:transferring}.
Finally, there is a notable effort on Arabic sign-language translation at King Fahd University of Petroleum and Minerals \cite{luqman2019automatic}.
For recent surveys of Arabic MT, see \cite{Ameur2020ArabicMT}. %,Harrat:2017:machine}.
Despite all these contributions, much research work is still needed to improve the performance of Machine translation for Arabic.

%%%%%%%%%%%%%%%%%%%%%%%%%%%%%%%%%%%%%%%%%%%%%%%%%%%%%%%%%%%%%%%%%%%%%%%%%%%%%%%%%%%%%%%%%%%%%%%%
%%%%%%%%%%%%%%%%%%%%%%%%%%%%%%%%%%%%%%%%%%%%%%%%%%%%%%%%%%%%%%%%%%%%%%%%%%%%%%%%%%%%%%%%%%%%%%%%
%%%%%%%%%%%%%%%%%%%%%%%%%%%%%%%%%%%%%%%%%%%%%%%%%%%%%%%%%%%%%%%%%%%%%%%%%%%%%%%%%%%%%%%%%%%%%%%%
\subsubsection{Pedagogical Applications} % Round 2 - initial; needs a survey citation/book AND organization to show what was done in the Arab World or outside it.  
% Violetta Cavalli-Sforza

Pedagogical applications (PA) focus on building tools to develop or model four major skills: reading, writing, listening, and speaking.  Arabic PA research has solely focused on MSA.
PA systems can be distinguished in terms of their target learners as first language (L1) or second (foreign) language (L2) systems.  This distinction can be problematic since, for Arabs, learning to read MSA is somewhat akin to reading a foreign tongue due to its lexical and syntactic divergence from native dialects.
We focus our Arabic PA discussion on (a) computer-assisted language learning (CALL) systems, (b) readability assessment, and (c) resource-building efforts.

\textbf{CALL systems} utilize NLP enabling technologies to assist language learners.  There has been a number of efforts in Arabic CALL exploring a range of resources and techniques. Examples include the use of
Arabic grammar and linguistic analysis rules to help learners identify and correct a variety of errors \cite{shaalan2006error,shaalan2005intelligent}; 
and multi-agent tutoring systems that simulate the instructor, the student, the learning strategy, and include a log book to monitor progress, and a learning interface \cite{mahmoud2018multiagents,mahmoud2016intelligent}.  
Another approach focuses on enriching the reading experience with concordances, text-to-speech (TTS), morpho-syntactic analysis, and auto-generated quiz questions \cite{maamouri2012developing}.

\textbf{Readability Assessment} is the task of automatic identification of a text's readability, i.e., its ability to be read and understood by its reader employing an acceptable amount of time and effort.
There has been a range of approaches for Arabic L1 and L2 readability. On one end, we find formulas using language-independent variables such as text length, average word length, and average sentence
length, number of syllables in words, the relative rarity or absence of dialectal alternatives, and the presence of less common letters \cite{c-s2018review,el-haj-rayson-2016-osman}. 
Others integrate Arabic morphological, lexical and syntactic features with supervised machine learning approaches \cite{alkhalifa2010arability,alTamimi2014AARI,saddiki-etal-2018-feature}.

%CHECK CITATIONS %No entry c-s2014elmezouar - is it this: cavalli2014matching,

Although some progress has been made for both L1 and L2 PA, the dearth of resources compared with English remains the bottleneck for future progress. \textbf{Resource-building} efforts have focused on L1 readers with particular emphasis on grade school curricula \cite{Al-Ajlan2008,alkhalifa2010arability,alTamimi2014AARI}.  
There is a push to inform the enhancement of curricula using pedagogical tools and to compare
curricula across Arab countries \cite{al-sulaiti-etal-2016-compilation,mubarak2020curricula}. 
The L2 PAs are even more limited with limited corpora \cite{maamouri2012developing} 
with disproportionate focus on beginners.\footnote{\url{https://learning.aljazeera.net/en}}
There is a definite need for augmenting these corpora in a reasoned way, taking into consideration different text features and learners, both young and old, beefing up the sparsely populated levels with authentic material, and exploiting technologies such as text simplification and text error analysis and correction. 
Learner corpora, which as the name suggests are produced by learners of Arabic can inform the creation of tools and corpora \cite{Alfaifi2014ArabicLC,rytting-etal-2014-arcade,Zaghouani:2015:correction}.
%No entry! alkanhal2012kacstc,alotaibi2017parallel
A recent effort developed a large-scale Arabic readability lexicon compatible with an existing morphological analysis system \cite{al-khalil-etal-2020-large}.

\subsubsection{Information Retrieval and Question Answering} % Kareem -- Not done | Hi Kareem - I wrapped this up.
% Hussein Al-Nathsheh & Kareem Darwish
With the increasing volume of Arabic content, information retrieval (aka search) has become a necessity for many domains such as medical records, digital libraries, web content, and news. The main research interests have focused on retrieval of formal language, mostly in the news domain, with ad hoc retrieval \cite{oard2002trec}, OCR document retrieval \cite{darwish2007error,darwish2002term}, and cross-language retrieval \cite{elayeb2016arabic}. The literature on other aspects of retrieval continues to be sparse or non-existent, though some of these aspects have been investigated by industry. Others aspects of Arabic retrieval that have received some attention include document image retrieval \cite{magdy2009efficient}, speech search \cite{olive2011handbook}, social media \cite{darwish2012language,hasanain2018evetar} and web search \cite{hasanain2020artest,suwaileh2016arabicweb16}, and filtering \cite{darwish2014arabic,magdy2016unsupervised}. However, efforts on different aspects of Arabic retrieval continue to be deficient and severely lacking behind efforts in other languages. Examples of unexplored problems include searching Wikipedia, which contains semi-structured content, religious text \cite{malhas2020ayatec}, which often contain semi-structured data such chains of narrations, rulings, and commentaries, Arabic forums, which are very popular in the Arab world and constitute a significant portion of the Arabic web, and poetry.  To properly develop algorithms and methods to retrieve such content, standard test sets and clear usage scenarios are required. We expect that recent improvements in contextual embeddings can positively impact the effectiveness of many retrieval tasks.

Another IR related problem is question answering, which comes in many flavors, the most common of which is attempting to identify a passage or a sentence that answers a question \cite{hasanain2014identification}.  Performing such a task may employ a large set of NLP tools such as parsing, NER, co-reference resolution, and text semantic representation. 
There has been limited research on this problem  \cite{IEEE:mawdoo3/deep2019,romeo2019language}
and existing commercial solutions such as \url{Ujeeb.com} are rudimentary.

\subsubsection{Dialogue Systems} 
% Violetta Cavalli-Sforza
% Karim Bouzoubaa
Automated dialog systems capable of sustaining a smooth and natural conversation with users have attracted considerable interest from both research and industry in the past few years. This technology is changing how companies engage with their customers among many other applications. While commercial dialog systems by big multinational companies such as Amazon Alexa, Google Home, and Apple Siri support many languages, only Apple Siri supports Arabic with limited performance.  There are some strong recent competitors in the Arab world, particularly Arabot\footnote{\url{https://arabot.io/}} and Mawdoo3's Salma.\footnote{\url{http://salma.ai/}}

While there is an important growing body of research on English language dialog systems,
current NLP methods for Arabic language dialogue are mostly based on handcrafted rule-based systems and methods that use feature engineering \cite{alhumoud2018arabic}. 
Among the earliest research efforts on Arabic dialog applications is the Quran chatbot \cite{shawar2004arabic},  
where the conversation length is short since the system answers a user input with a single response. It uses a retrieval-based model as the dataset is limited by the content of the Quran. A recent approach used deep learning techniques for text classification and NER to build a natural language understanding module -- the core component of any dialogue system -- for the domain of home automation in Arabic \cite{bashir2018implementation}. 
A unique dialogue system from NYUAD explored bilingual interfaces where Arabic speech can be used as input to an English bot that displays Arabic subtitles \cite{abu-ali-etal-2018-bilingual}.
Other works have focused on developing dialog systems for the case of Arabic dialects, e.g. 
the publicly available NYUAD Egyptian dialect chatbot {\it Botta},
%\cite{ali2016botta},
and KSU's Saudi dialect Information Technology focused chatbot {\it Nabiha} \cite{alnabiha}.

\subsubsection{Sentiment and Emotion Analysis} % Round 2 - DONE - Nizar+Kareem 
% Samhaa El-Beltagy
% Wassim El-Hajj

%What
Sentiment analysis (SA), aka opinion mining, is the task of identifying the
affective states and subjective information in a text. For example, an Egyptian Arabic
movie review such as <A.hsn fylm alsnT dy!> `the best movie this
year!' is said to indicate a positive sentiment.
%Why
SA is a very powerful tool for tracking customer satisfaction, carrying out competition analysis,
and generally gauging public opinion towards a specific
issue, topic, or product.
%in the Arab World
%
SA has attracted a lot of attention in the Arabic research community during the
last decade, connected with the availability of large volumes of opinionated and
sentiment reflecting data from  Arabic social media.
%Resources and Benchmarks/Shared Tasks
Early Arabic SA efforts focused on the creation of needed resources
such as sentiment lexicons, training datasets, and sentiment treebanks \cite{Abdul-Mageed2012,Badaro:2014:large,baly2017sentiment,El-Beltagy2016,ElSahar2015,Eskander2015,Khalil2015,Mourad2013,Refaee2014,salameh-mohammad-kiritchenko:2015:NAACL-HLT,Shoukry2015}, 
as well as shared task benchmarks \cite{SemEval2018Task1,SemEval:2017:task4}.
%
%Solutions
Arabic SA solutions span a range of methods from the now conventional use of
rules and lexicons \cite{Abdul-Mageed:2014:sana,Badaro:2014:large,El-Beltagy2013}
to machine learning based methods \cite{abu-farha-magdy-2019-mazajak,badaro-etal-2018-ema,baly2017sentiment},
as well as hybrid approaches employing morphological and syntactic features \cite{al2019enhancing}. 
Recently, fine-tuning large pre-trained language models has achieved improved Arabic SA results \cite{abdulmageed2020microdialect,antoun2020arabert}.
Arabic emotion recognition is a closely related topic that has attracted some attention recently. It aims to identify a variety of emotions in text such as anger, disgust, surprise, and joy \cite{alswaidan2020hybrid,badaro2018arsel,baly2019arsentd,hifny2019efficient,shahin2019emotion}. 
Similar to how SA resources and models started maturing, a lot of work still needs to be done in emotion recognition. Another related problem is stance detection, which attempts to identify positions expressed on a topic or towards an entity.  Stances are often expressed using non-sentiment words \cite{borge2015content}.
For a recent comprehensive survey on the status of Arabic SA and the future directions, see \cite{badaro2019survey}.

\subsubsection{Content Moderation on Social Media}% Round 2 - done.
% Hamdy Mubarak
The task of content moderation is about the enforcement of online outlets' policies against posting user comments that contain offensive language, hate speech, cyber-bullying, and spam among other types of inappropriate or dangerous  content.\footnote{\url{https://www.bbc.co.uk/usingthebbc/terms/what-are-the-rules-for-commenting/}}
Such content cannot be easily detected given the huge volume of posts, dialectal variations, creative spelling on social media, and the scarcity of available data and detection tools. 
This area is relatively new for Arabic.  One of the more active areas has to do with the detection of offensive language, which covers  targeted attacks, vulgar and pornographic language, and hate speech. Initial work was performed on comments from a news site and limited numbers of tweets \cite{mubarak2017abusive}
and YouTube comments \cite{alakrot2018towards}. 
Some works focused on adult content \cite{Alshehri2018ThinkBY} and others on hate speech.
%\cite{albadi2018they,al2019detection}. 
Recent benchmarking shared tasks included the automatic detection of such language in Twitter domain \cite{mubarak-etal-2020-overview}. Work on spam detection on Twitter is nascent and much work is required \cite{mubarak2020spam}.

\section{Future Outlook} % Round 2 - done.
Arabic NLP has many challenges, but it has also seen many successes and developments over the last 40 years. We are optimistic by its continuously positive albeit (sometimes) slow development trajectory.  
For the next decade or two, we expect a large growth in the Arabic NLP market. This is consistent with the global rising demands and expectations for language technologies and the increase in NLP research and development in the Arab world.  The growing number of researchers and developers working on NLP in the Arab world makes it a very fertile ground ready for major breakthroughs. 
%
%While ML techniques are here to stay and improve, we expect a continuing role for morphological analyzers and dictionaries -- explainable AI?
%
To support this vision, we believe it is time to have an association for Arabic language technologists that brings together talent and resources, and sets standards for the Arabic NLP community.  Such an organization can support NLP education in the Arab world, serve as a hub for resources, and advocate for educators and researchers in changing old-fashioned university policies regarding journal-focused evaluation, and encouraging collaborations within the Arab world by connecting academic, industry, and governmental stakeholders. We also recommend more open source tools and public data are made available to create a basic development framework that lowers the threshold for joining the community, thus attracting more talent that will form the base of the next generation of Arabic NLP researchers, developers, and entrepreneurs.

%\begin{itemize}
%    \item  An association for Arabic Language Technologies?
%    \item  A hub for data and resources; open
%    \item   NLP education in the Arab world
%    \item   Attracting Arab talent to work on Arabic 
%    \item   More basic common frameworks for development
%\end{itemize}

%\bibliography{ref} 
\bibliography{ref}
\bibliographystyle{acm}

\end{document}